\documentclass[12pt,onecolumn]{IEEEtran}
\usepackage{tikz}

\usepackage{times}
\usepackage{amsmath}  
\usepackage{amsfonts}
\usepackage{amsthm}

\usepackage{epsfig}
\usepackage{graphicx}
\usepackage{epstopdf}

\renewcommand{\eqref}[1]{(\ref{#1})}

\newcommand{\var}{\ensuremath{\mathrm{Var}}}
\newcommand{\E}{\ensuremath{\mathbb{E}}}
\renewcommand{\P}{\ensuremath{\mathbb{P}}}
\newcommand{\R}{\ensuremath{\mathbb{R}}}
\newcommand{\Z}{\ensuremath{\mathbb{Z}}}

\def\mvo{mean-variance optimization}
\def\Mvo{Mean-variance optimization}
\def\mv{{\sc mv-mdp}}
\def\mS{{\cal S}}
\def\mA{{\cal A}}
\def\mR{{\cal R}}
\def\mM{{\cal M}}
\def\mX{{\cal X}}
\def\Jp{J_{\pi}}
\def\Vp{V_{\pi}}
\def\Qp{Q_{\pi}}
\def\smskip{\par\vspace{7pt}}
\def\qed{\hspace{5pt}{\bf q.e.d.}}
\def\ox{{\overline x}}

\def\zp{z^{\pi}}

\newcommand{\conv}{\mathop{\rm conv}}

\newtheorem{theorem}{Theorem}
\newtheorem{proposition}{Proposition}

\title{Mean-Variance Optimization in Markov Decision Processes}

\author{Shie Mannor \\ (Corresponding Author)  Department of Electrical and Engineering, Technion, \\ Haifa, ISRAEL 32000, tel ++972-4-8293284, fax ++972-4-8295757 \\
\and John N. Tsitsiklis \\ Laboratory for Information and Decision Systems, \\Massachusetts
Institute of Technology, Cambridge, MA, 02139}

\usepackage{apacite}

\begin{document}

\maketitle

\begin{abstract}
\noindent
We consider finite horizon Markov decision processes under performance measures that involve both the mean and the variance of the cumulative reward. We show that either randomized or history-based policies 
can improve performance. We prove that the complexity of computing a policy that maximizes the mean reward under a variance constraint is NP-hard for some cases, and strongly NP-hard for others. We finally offer pseudopolynomial exact and approximation algorithms.
\end{abstract}

\baselineskip=24pt

\noindent
{\bf keywords}: Markov processes; dynamic programming; control; complexity theory.

\section{Introduction}
The classical theory of Markov decision processes (MDPs) deals with the maximization of the cumulative (possibly discounted) expected reward, to be denoted by $W$.
However, a risk-averse decision maker  may be interested in additional distributional properties of $W$. In this paper, we focus on the case where the decision maker is interested in both the mean and the variance of the cumulative reward, and we explore the associated computational issues.

Risk aversion in MDPs is of course an old subject. In one approach, the focus is on the maximization of $\E[U(W)]$, where $U$ is a concave utility function. Problems of this type can be handled by state augmentation \cite<e.g.,>{Bertsekas95}, namely, by introducing an auxiliary state variable that keeps track of the cumulative past reward. In a few special cases, e.g., with an exponential utility function, state augmentation is unnecessary, and optimal policies can be found by solving a modified Bellman equation \cite{ChungSobel87:Discountedexponential}. Another interesting case where optimal policies can be found efficiently involves 
piecewise linear utility  functions with a single break point; see \citeA{LiuKoenig05:risksensitiveoneswitch}.

In another approach, the objective is to optimize a so-called coherent risk measure \cite{coherent}, which turns out to be equivalent to a  robust optimization problem: one assumes a family of probabilistic models and optimizes the worst-case performance over this family. In the multistage case \cite{coherentdp}, problems of this type can be difficult \cite{yann}, except for some special cases \cite{iyengar05:rdp,nilim05:rmdputm} that can be reduced to Markov games \cite{S53}.

\Mvo\ lacks some of the desirable properties of approaches involving coherent risk measures and sometimes leads to counterintuitive policies. Bellman's principle of optimality  does not hold, and as a consequence, a decision maker who has received unexpectedly large rewards in the first stages, may actively seek to incur losses in subsequent stages in order to keep the variance small. Nevertheless, \mvo\ is an important approach in financial decision making \cite<e.g.,>{luenberger}, especially for
static (one-stage) problems. Consider, for example, a fund manager who is interested in the 1-year performance of the fund, as measured by the mean and variance of the return. Assuming that the manager is allowed to undertake periodic re-balancing actions in the course of the year, one obtains a Markov decision process with mean-variance criteria.
\Mvo\ can also be a meaningful objective in various engineering contexts.
Consider, for example, an engineering process whereby a certain material is deposited on a surface. Suppose that the primary objective is to maximize the amount deposited, but that there is also an interest in having all manufactured components be similar to each other; this secondary objective can be addressed by keeping the variance of the amount deposited small.

We note that expressions for the variance of the discounted reward for stationary policies were developed in \citeA{Sobel82variance}. However, these expressions 
are quadratic in the underlying transition probabilities, and do not lead to convex optimization problems. 

Motivated by considerations such as the above, this paper deals with the computational complexity aspects of \mvo. The problem is not straightforward for various reasons. One is the absence of a principle of optimality that could lead to simple recursive algorithms. Another reason is that, as is evident from the formula $\var(W)=\E[W^2]-(\E[W])^2$, the variance is not a linear function of the probability measure of the underlying process. Nevertheless, $\E[W^2]$ and $\E[W]$ are linear functions, and as such can be addressed simultaneously using methods from multicriteria or constrained Markov decision processes \cite{Alt99}. Indeed, we will use such an approach in order to develop pseudopolynomial exact or approximation algorithms. On the other hand, we will also obtain various NP-hardness results, which show that there is little hope for significant improvement of our algorithms.

The rest of the paper is organized as follows. In Section \ref{s:model}, we describe the model and our notation. We also define various classes of policies and performance objectives of interest. In Section \ref{s:comp}, we compare different policy classes and show that performance typically improves strictly as more general policies are allowed. In Section \ref{s:compl}, we establish NP-hardness results for the policy classes we have introduced. Then, in Sections \ref{s:algs} and \ref{se:approx}, we develop exact and approximate pseudopolynomial time algorithms.
Unfortunately, such algorithms do not seem possible for some of the more restricted classes of policies, due to  strong NP-completeness results established in Section \ref{s:compl}. 
Finally, Section \ref{s:concl} contains some brief concluding remarks.

\section{The Model}
\label{s:model}
In this section, we define the model, notation, and performance objectives that we will be studying. Throughout, we focus on finite horizon problems. \footnote{Some of the results such as the approximation algorithms of Section 
\ref{se:approx} can be extended to the infinite horizon discounted case; this is beyond the scope of this paper.}

\subsection{Markov Decision Processes}\label{s:mdps}

We consider a Markov decision process (MDP) with finite state, 
action, and reward spaces. An MDP is formally defined by a sextuple
$\mM=(T,\mS,\mA,\mR,p,g)$ where:
\begin{enumerate}
\item[(a)] $T$, a positive integer, is the time horizon;
\item[(b)]
$\mS$ is a finite collection of states, one of which is designated as the initial state;
\item[(c)] $\mA$ is a collection of finite sets of possible actions, one set for each state;
\item[(d)] $\mR$ is a finite subset of $\mathbb{Q}$ (the set of rational numbers), and is the set of possible values of the immediate rewards.
We let $K=\max_{r\in\mR}|r|$.
\item[(e)] $p:\{0,\ldots,T-1\}\times \mS\times \mS\times \mA\to \mathbb{Q}$ describes the transition probabilities. In particular, $p_t(s'\,|\,s,a)$
is the probability that the state at time $t+1$ is $s'$, given that
the state at time $t$ is $s$, and that action $a$ is chosen at time $t$.
\item[(d)] $g:\{0,\ldots,T-1\}\times \mR\times\mS\times\mA\to\mathbb{Q}$ is a set of reward distributions. In particular, $g_t(r\,|\,s,a)$ is
the probability that the immediate reward at time $t$ is $r$, given that the state and action at time $t$ is $s$ and $a$, respectively.
\end{enumerate}
With few exceptions (e.g., for the time horizon $T$), we use capital letters to denote random variables, and lower case letters to denote ordinary variables.
The process starts at the designated initial state.
At every stage $t=0,1,\ldots,T-1$, the decision maker observes the current
state $S_t$ and chooses an action $A_t$. Then,
an immediate reward $R_t$ is obtained, distributed according to $g_t(\,\cdot\,|\,S_t,A_t)$,
and the next state
$S_{t+1}$ is chosen, according to $p_t(\,\cdot\,|\,S_t,A_t)$.
Note that we have assumed that the possible values of the immediate reward and the various probabilities are all rational numbers. This is in order to address the computational complexity of various problems within the standard framework of digital computation.
Finally, we will use the notation $x_{0:t}$ to indicate the tuple $(x_0,\ldots,x_t)$.

\subsection{Policies}
We will use the symbol $\pi$ to denote policies. Under a {\it deterministic policy} $\pi=(\mu_0,\ldots,\mu_{T-1})$, the action at each time $t$ is determined according to a mapping $\mu_t$ whose argument is the history $H_t=(S_{0:t},A_{0:t-1},R_{0:t-1})$ of the process, by letting $A_t=\mu_t
(H_t)$.
We let $\Pi_h$ be the set of all such history-based policies.
(The subscripts are used as a mnemonic for the variables on which the action is allowed to depend.)
We will also consider {\it randomized} policies. For this purpose, we assume that there is available a sequence of i.i.d.\ uniform random variables $U_0,U_1,\ldots,U_{T-1}$, which are independent from everything else. In a randomized policy, the action at time $t$ is determined by letting $A_t=\mu_t(H_t,U_{0:t})$. Let $\Pi_{h,u}$ be the set of all randomized policies.

In classical MDPs, it is well known that restricting to Markovian policies (policies that take into account only the current state $S_t$) results in no loss of performance. 
In our setting, there are two different possible ``states'' of interest: the original state $S_t$, or the augmented state $(S_t,W_t)$, where
$$W_t=\sum_{k=0}^{t-1}R_k,$$
(with the convention that $W_0=0$). Accordingly, we define the following classes of policies: $\Pi_{t,s}$ (under which $A_t=\mu_t(S_t)$), and $\Pi_{t,s,w}$ (under which $A_t=\mu_t(S_t,W_t)$), and their randomized counterparts $\Pi_{t,s,u}$ (under which $A_t=\mu_t(S_t,U_t)$), and $\Pi_{t,s,w,u}$ (under which $A_t=\mu_t(S_t,W_t,U_t)$. Notice that
$$\Pi_{t,s} \subset \Pi_{t,s,w} \subset \Pi_{h},$$
and similarly for their randomized counterparts.

\subsection{Performance Criteria}

Once a policy $\pi$ and an initial state $s$ is fixed, the cumulative reward $W_T$ becomes a well-defined random variable. The performance measures of interest are its mean and variance, defined by $J_{\pi}=\E_{\pi}[W_T]$ and
$V_{\pi}=\var_{\pi}(W_T)$, respectively. Under our assumptions (finite horizon, and bounded rewards), it follows that
there are finite upper boundsof $KT$ and $K^2 T^2$, for $|J_{\pi}|$ and $V_{\pi}$, respectively, independent of the policy.

Given our interest in complexity results, we will focus on ``decision'' problems that admit a yes/no answer, except for Section \ref{se:approx}. We define the following problem.

\smskip
\noindent
{\bf Problem \mv($\Pi$):} Given an MDP $\mM$ and rational numbers $\lambda$, $v$, 
does there exist a policy in the set $\Pi$ such that $\Jp\geq \lambda$ and $\Vp\leq v$?
\smskip

Clearly, an algorithm for the problem \mv($\Pi$) can be combined with binary search to solve (up to any desired precision) the problem of maximizing the expected value of $W_T$ subject to an upper bound on its variance,
or the problem of minimizing the variance of $W_T$ subject to a lower bound on its mean.

\section{Comparison of Policy Classes}
\label{s:comp}

Our first step is to compare the performance obtained from different policy classes. We introduce some terminology. Let $\Pi$ and $\Pi'$ be two policy classes.
We say that $\Pi$ is {\it inferior} to $\Pi'$ if, loosely speaking, the policy class $\Pi'$ can always match or exceed the ``performance'' of policy class $\Pi$, and for some instances it can exceed it strictly. Formally, $\Pi$ is inferior to $\Pi'$ if the following hold: (i) if $(\mM,c,d)$ is a ``yes'' instance of \mv($\Pi$), then it is also a ``yes'' instance of \mv($\Pi'$); (ii) there exists some
$(\mM,c,d)$ which is a ``no'' instance of \mv($\Pi$) but a ``yes'' instance of
\mv($\Pi'$).
Similarly, we say that two policy classes $\Pi$ and $\Pi'$ are {\it equivalent} if every ``yes'' (respectively, ``no'') instance of \mv($\Pi$) is a ``yes'' (respectively, ``no'') instance of \mv($\Pi'$).

We define one more convenient term. A state $s$ is said to be  {\it terminal} if it is absorbing (i.e., $p_t(s\,|\,s,a)=1$, for every $t$ and $a$) and provides zero rewards (i.e., $g_t(0\,|\,s,a)=1$, for every $t$ and $a$).

\subsection{Randomization Improves Performance}

Our first observation is that randomization can improve performance. This is not surprising given that we are dealing  simultaneously with two criteria, and that randomization is helpful in constrained MDPs \cite<e.g.,>{Alt99}.

\begin{theorem} \label{th:random}
\begin{itemize}
\item[(a)] $\Pi_{t,s}$ is inferior to $\Pi_{t,s,u}$;
\item[(b)] $\Pi_{t,s,w}$ is inferior to $\Pi_{t,s,w,u}$;
\item[(c)] $\Pi_{h}$ is inferior to $\Pi_{h,u}$.
\end{itemize}
\end{theorem}
\noindent
{\bf Proof.}  It is clear that performance cannot deteriorate when randomization is allowed. It therefore suffices to display an instance in which randomization improves performance.

Consider a one-stage MDP ($T=1$). At time $0$, we are at the initial state and there are two available actions, $a$ and $b$. The mean and variance of the resulting reward are both zero under action $a$, and both equal to 1 under action $b$. After the decision is made, the rewards are obtained and the process terminates. Thus $W_T=R_0$, the reward obtained at time 0.

Consider the problem of maximizing $\E[R_0]$ subject to the constraint that
$\var(R_0)\leq 1/2$. There is only one feasible deterministic policy (choose  action $a$), and it has zero expected reward. On the other hand, a randomized policy that chooses action $b$ with probability $p$ has an expected reward of $p$ and the corresponding variance satisfies
$$\var(R_0)\leq \E[R_0^2] =p \E[R_0^2\,|\, A_0=b]=2p.$$
When $0<p\leq 1/4$, such a randomized policy is feasible and improves upon the deterministic one.

Note that for the above instance we have
$\Pi_{t,s}=\Pi_{t,s,w}=\Pi_{h}$, and $\Pi_{t,s,u}= \Pi_{t,s,w,u}=\Pi_{h,u}$. Hence the above example establishes all three of the claimed statements.
\qed

\subsection{Information Improves Performance}

We now show that in most cases, performance can improve strictly when we allow a policy to have access to more information. The only exception arises for the pair of classes $\Pi_{t,s,w,u}$ and $\Pi_{h,u}$, which we show in Section \ref{s:algs} to be equivalent (cf.\ Theorem \ref{th:equiv}).

\begin{theorem} \label{th:ranking}
\begin{itemize}
\item[(a)] $\Pi_{t,s}$ is inferior to $\Pi_{t,s,w}$, and
$\Pi_{t,s,u}$ is inferior to $\Pi_{t,s,w,u}$.
\item[(b)] $\Pi_{t,s,w}$ is inferior to $\Pi_{h}$.
\end{itemize}
\end{theorem}
\noindent
{\bf Proof.}
\begin{itemize}
\item[(a)]
Consider the following MDP, with time horizon $T=2$. The process starts at the initial state $s_0$, at which there are two actions. Under action $a_1$, the immediate reward is zero and the process moves to a terminal state. Under action $a_2$, the immediate reward $R_0$ is either 0 or 1, with equal probability, and the process moves to state $s_1$. At state $s_1$, there are two actions, $a_3$ and $a_4$: under action $a_3$, the immediate reward $R_1$ is equal to 0, and under action $a_4$, it is equal to 1. We are interested in the optimal value of the expected reward $\E[W_2]=\E[R_0+R_1]$, subject to the constraint that the variance is less than or equal to zero (and therefore equal to zero). Let $p$ be the probability that action $a_2$ is chosen at state $s_0$. If $p>0$, and under any policy in $\Pi_{t,s,u}$, the reward $R_0$  at state $s_0$ has positive variance, and the reward $R_1$ at the next stage is uncorrelated with $R_0$. Hence, the variance of $R_0+R_1$ is positive, and such a policy is not feasible; in particular, the constraint on the variance requires that $p=0$. We conclude that the largest possible expected reward under any policy in $\Pi_{t,s,u}$ (and, a fortiori, under any policy in $\Pi_{t,s}$) is equal to zero.

Consider now the following policy, which belongs to $\Pi_{t,s,w}$ and, a fortiori, to $\Pi_{t,s,w,u}$: at state $s_0$, choose action $a_2$; then, at state $s_1$, choose $a_3$ if $W_1=R_0=1$, and choose $a_4$ if $W_1=R_0=0$. In either case, the total reward is $R_0+R_1=1$, while the variance of $R_0+R_1$ is zero, thus ensuring feasibility. This establishes the first part of the theorem.

\item[(b)] Consider the following MDP, with time horizon $T=3$. At state $s_0$ there is only one available action; the next state $S_1$ is either $s_1$ or $s_1'$, with probability $p$ and $1-p$, respectively, and the immediate reward $R_0$ is zero. At either state $s_1$ or $s_1'$, there is again only one available action; the next state, $S_2$, is $s_2$, and the reward $R_1$ is zero. At state $s_2$, there are two actions, $a$ and $b$. Under action $a$, the mean and variance of the resulting reward $R_2$ are both zero, and under action $b$, they are both equal to 1. Let us examine the largest possible value of $\E[W_3]=\E[R_2]$,
subject to the constraint $\var(W_2)\leq 1/2$. The class $\Pi_{t,s,w}$ contains two policies, corresponding to the two deterministic choices of an action at state $s_2$; only one of them is feasible (the one that chooses action $a$), resulting in zero expected reward. However, the following policy in $\Pi_h$ has positive expected reward: choose action $b$ at state $s_2$ if and only if the state at time 1 was equal to $s_1$ (which happens with probability $p$). As long as $p$ is sufficiently small, the constraint $\var(W)\leq 1/2$ is met, and this policy is feasible. It follows that $\Pi_{t,s,w}$ is inferior to $\Pi_h$.\qed
\end{itemize}

\section{Complexity Results}
\label{s:compl}

In this section, we establish that mean-variance optimization in finite horizon MDPs is unlikely to admit polynomial time algorithms, in contrast to classical MDPs.

\begin{theorem} \label{th:wnpc}
The problem \mv($\Pi$) is NP-hard, when $\Pi$ is
$\Pi_{t,s,w}$, $\Pi_{t,s,w,u}$, $\Pi_h$,  or $\Pi_{h,u}$.
\end{theorem}
\noindent
{\bf Proof:}
We will actually show NP-hardness for the special case of \mv($\Pi$), in which we wish to determine whether there exists a policy whose reward variance is equal to zero. (In terms of the problem definition, this corresponds to letting 
$\lambda=-KT$ and $v=0$.)
The proof uses a reduction from the {\sc subset sum} problem:
Given $n$ positive integers, does there exist a subset $B$ of $\{1,\ldots,n\}$ such that $\sum_{i\in B} r_i =\sum_{i\notin B} r_i$?

Given an instance $(r_1,\ldots,r_n)$ of {\sc subset sum}, and for any of the policy classes of interest, we construct an instance of \mv($\Pi$), with time horizon $T=n+1$, as follows. At the initial state $s_0$, there is only one available action, resulting in zero immediate reward ($R_0=0$). With probability 1/2, the process moves to a terminal state; with probability 1/2, the process moves (deterministically) along a sequence of states $s_1,\ldots,s_n$.  At each state $s_i$ ($i=1,\ldots,n$), there are two actions: $a_i$, which results in an immediate reward of $r_i$, and $b_i$, which results in an immediate reward of $-r_i$.

Suppose that there exists a set $B\subset\{1,\ldots,n\}$ such that
$\sum_{i\in B} r_i =\sum_{i\notin B} r_i$. Consider the policy that chooses action $a_i$ at state $s_i$ if and only if $i\in B$. This policy achieves zero total reward, with probability 1, and therefore meets the zero variance constraint. Conversely, if a policy results in zero variance, then the total reward must be equal to zero, with probability 1, which implies that such a set $B$ exists. This completes the reduction.

Note that this argument applies no matter which particular class of policies is being considered.
\qed

The above proof also applies to the policy classes $\Pi_{t,s}$ and $\Pi_{t,s,u}$. However, for these two classes, a stronger result is possible. Recall that a problem is {\it strongly NP-hard}, if it remains NP-hard when restricted to instances in which the numerical part of the instance description involves ``small'' numbers; see \citeA{GareyJohnson79:GuidetoNPC} for a precise definition.

\begin{theorem}\label{th:snpc}
If\/ $\Pi$ is either $\Pi_{t,s}$ or $\Pi_{t,s,u}$, the problem \mv($\Pi$) is strongly NP-hard.
\end{theorem}
\noindent
{\bf Proof.} As in the proof of Theorem \ref{th:wnpc},
we will prove the result for the special case of \mv, in which we wish to determine whether there exists a policy under which the variance of the reward is equal to zero.
The proof involves a reduction from the 3-Satisfiability problem ({\sc 3sat}). An instance of {\sc 3sat} consists of $n$ Boolean variables $x_1,\ldots,x_n$, and $m$ clauses $C_1,\ldots,C_m$, with three literals per clause. Each clause is the disjunction of three literals, where a literal is either a variable or its negation. (For example, $x_2\vee \ox_4 \vee x_5$ is such a clause, where a bar stands for negation.) The question is whether there exists an assignment of truth values (``true'' or ``false'') to the variables such that all clauses are satisfied.

Suppose that we are given an instance of {\sc 3sat}, with $n$ variables and $m$ clauses, $C_1,\ldots,C_m$. We construct an instance of \mv($\Pi$) as follows.
There is an initial state $s_0$, a state $d_0$, a state $c_j$ associated with each clause $C_j$, and a state $y_i$ associated with each literal $x_i$. The actions, dynamics, and rewards are as follows:
\begin{itemize}
\item[(a)] Out of state $s_0$, there is equal probability, $1/(m+1)$, of reaching  any one of the states $d_0,c_1,\ldots,c_m$, independent of the action; the immediate reward is zero.
\item[(b)] State $d_0$ is a terminal state. At each state $c_j$, there are three actions available: each action selects one of the three literals in the clause, and the process moves to the state $y_i$ associated with that literal; the immediate reward is 1 if the literal appears in the clause unnegated, and $-1$ if the literal appears in the clause negated. For an example, suppose that the clause is of the form $x_2\vee \ox_4\vee x_5$. Under the first action, the next state is $y_2$, and the reward is 1; under the second action, the next state is $y_4$ and the reward is $-1$; under the third action, the next state is $y_5$, and the reward is 1.
\item[(c)] At each state $y_i$, there are two possible actions $a_i$ and $b_i$, resulting in immediate rewards of 1 and $-1$, respectively. The process then moves to the terminal state $d_0$.
\end{itemize}

Suppose that we have a ``yes'' instance of {\sc 3sat}, and consider a truth assignment that satisfies all clauses. We can then construct a policy in $\Pi_{t,s}$ (and a fortiori in $\Pi_{t,s,u}$, whose total reward is zero (and therefore has zero variance) as follows. If $x_i$ is set to be true (respectively, false), we choose action $b_i$ (respectively, $a_i$) at state $y_i$. At state $c_j$ we choose an action associated with a literal that makes the clause to be true. Suppose that state $c_j$ is visited after the first transition, i.e., $S_1=c_j$. If the literal associated with the selected action at $c_j$ is unnegated, e.g., the literal $x_i$, then the immediate reward is 1. Since this literal makes the clause to be true, it follows that the action chosen at the subsequent state, $y_i$, is $b_i$, resulting in a reward of $-1$, and a total reward of zero. The argument for the case where the literal associated with the selected action at state $c_j$ is negated is similar. It follows that the total reward is zero, with probability 1.

For the converse direction, suppose that there exists a policy in $\Pi_{t,s}$, or more generally, in $\Pi_{t,s,u}$ under which the variance of the total reward is zero. Since the total reward is equal to 0 whenever the first transition leads to state $d_0$ (which happens
with probability $1/(m+1)$, it follows that the total reward must be always zero. Consider now the following truth assignment: $x_i$ is set to be true if and only if the policy chooses action $b_i$ at state $y_i$, with positive probability. Suppose that the state visited after the first transition is $c_j$. Suppose that the action chosen at state $c_j$ leads next to state $y_i$ and that the literal $x_i$ appears unnegated in clause $C_j$. Then, the reward at state $c_j$ is 1, which implies that the reward at state $y_i$ is $-1$. It follows that the action chosen at $y_i$ is $b_i$, and therefore $x_i$ has been set to be true. It follows that clause $C_j$ is satisfied. A similar argument shows that clause $C_j$ is satisfied when the literal $x_i$ associated with the chosen action at $c_j$ appears negated. In either case, we conclude that clause $C_j$ is satisfied. Since every state $c_j$ is possible at time 1, it follows that every clause is satisfied, and we have a ``yes'' instance of {\sc 3sat}.
\qed

\section{Exact Algorithms}
\label{s:algs}
The comparison and complexity results of the preceding two sections indicate that the 
policy classes $\Pi_{t,s}$, $\Pi_{t,s,w}$, $\Pi_{t,s,u}$, and $\Pi_h$ are inferior to the class $\Pi_{h,u}$, and furthermore some of them ($\Pi_{t,s}$, $\Pi_{t,s,w}$) appear to have higher complexity. Thus, there is no reason to consider them further. While the problem \mv($\Pi_{h,u}$) is NP-hard, there is still a possibility for approximate or pseudopolynomial time algorithms. In this section, we focus on exact pseudopolynomial time algorithms.

Our approach involves an augmented state, defined by
$X_t=(S_t,W_t)$. Let $\mX$ be the set of all possible values of the augmented state. Let $|\mS|$ be the cardinality of the set $\mS$. 
Let $|\mR|$ be the cardinality of the set $\mR$.
Recall also that $K=\max_{r\in\mR}|r|$.
If we assume that the immediate rewards are integers, then $W_t$ is an integer between $-KT$ and $KT$. In this case, the cardinality $|\mX|$ of the augmented state space $\mX$ is bounded by $|\mS|\cdot (2KT+1)$, which is polynomial. Without the integrality assumption, the cardinality of the set $\mX$ remains finite, but it can increase exponentially with $T$. For this reason, we study the integer case separately in Section \ref{s:integer}.

\subsection{State-Action Frequencies}
\label{s:sa}

In this section, we provide some results on the representation of MDPs in terms of a state-action frequency polytope, thus setting the stage for our subsequent algorithms.

For any policy $\pi\in\Pi_{h,u}$, and any $x\in\mX$, $a\in\mA$, we define the state-action frequencies at time $t$ by
$$\zp_t(x,a)=\P_{\pi}(X_t=x,A_t=a),\qquad t=0,1,\ldots,T-1,$$
and
$$\zp_t(x)=\P_{\pi}(X_t=x),\qquad t=0,1,\ldots,T.$$
Let $\zp$ be a vector that lists all of the above defined state-action frequencies.

For any family $\Pi$ of policies, let $Z(\Pi)=\{\zp\,|\, \pi\in\Pi\}$. The following result is well known \cite<e.g.,>{Alt99}. It asserts that any feasible state-action frequency vector can be attained by policies that depend only on time, the (augmented) state, and a randomization variable. Furthermore, the set of feasible state-action frequency vectors is a polyhedron, hence amenable to linear programming methods.

\begin{theorem}\label{th:polyh}
\begin{itemize}
\item[(a)] We have $Z(\Pi_{h,u})=Z(\Pi_{t,s,w,u})$.
\item[(b)] The set $Z(\Pi_{h,u})$ is a polyhedron, specified by $O(T\cdot |\mX| \cdot |\mA|)$ linear constraints.
\end{itemize}
\end{theorem}

Note that a certain mean-variance pair $(\lambda,v)$ is attainable by a policy in $\Pi_{h,u}$ if and only if there exists some $z\in Z(\Pi_{h,u})$ that satisfies
\begin{eqnarray}
\sum_{(s,w)\in\mX} w z_T(s,w)& =& \lambda, \label{eq:a}\\
\sum_{(s,w)\in\mX} w^2 z_T(s,w)& =& v+\lambda^2. \label{eq:b}
\end{eqnarray}
Furthermore, since $Z(\Pi_{h,u})=Z(\Pi_{t,s,w,u})$, it follows that if a pair $(\lambda,v)$ is attainable by a policy in $\Pi_{h,u}$, it is also attainable by a policy in $\Pi_{t,s,w,u}$. This establishes the following result.

\begin{theorem} \label{th:equiv}
The policy classes $\Pi_{h,u}$ and $\Pi_{t,s,w,u}$ are equivalent.
\end{theorem}

Note that checking the feasibility of the conditions $z\in Z(\Pi_{h,u})$, \eqref{eq:a}, and \eqref{eq:b} amounts to solving a linear programming problem, with a number of constraints proportional to the cardinality of 
the augmented state space $\mX$ and, therefore, 
in general, exponential in $T$.

\subsection{Integer Rewards}
\label{s:integer}
In this section, we assume that the immediate rewards are integers, with absolute value bounded by $K$, and we show that pseudopolynomial time  algorithms are possible. 
Recall that an algorithm is a
pseudopolynomial time algorithm if its running time is polynomial in $K$ and the instance size. (This is in contrast to polynomial time algorithms in which the running time can only grow as a polynomial of $\log K$.)

\begin{theorem} \label{th:pss}
Suppose that the immediate rewards are integers, with absolute value bounded by $K$. Consider the following two problems:
\begin{itemize}
\item[(i)] determine  whether there exists a policy in $\Pi_{h,u}$ for which $(\Jp,\Vp)=(\lambda,v)$, where $\lambda$ and $v$ are given rational numbers; and,
\item[(ii)] determine  whether there exists a policy in $\Pi_{h,u}$ for which $\Jp=\lambda$ and $\Vp\leq v$, where $\lambda$ and $v$ are given rational numbers.
\end{itemize}
Then,
\begin{itemize}
\item[(a)] these two problems admit a pseudopolynomial time algorithm; and, 
\item[(b)] unless P=NP, these problems cannot be solved in polynomial time.
\end{itemize}
\end{theorem}
\noindent
{\bf Proof.}
\begin{itemize}
\item[(a)] As already discussed, these problems amount to solving a linear program. In the integer case, the number of variables and constraints is bounded by a polynomial in $K$ and the instance size.
The result follows because linear programming can be solved in polynomial time.
\item[(b)] This is proved by considering the special case where $\lambda=v=0$ and the exact same argument as in the proof of Theorem \ref{th:wnpc}. \qed
\end{itemize}


Similar to constrained MDPs, \mvo\ involves two different performance criteria. Unfortunately, however, the linear programming approach to constrained MDPs does not translate into an algorithm for the problem \mv($\Pi_{h,u}$). The reason is that 
the set 
$$P_{MV}=\{(\Jp,\Vp)\mid \pi\in\Pi_{h,u}\}$$ 
of achievable mean-variance pairs need not be convex.
To bring the constrained MDP methodology to bear on our problem, instead of focusing on the pair $(\Jp,\Vp)$, we define $\Qp=\E_{\pi}[W^2_T]$, and focus on the pair $(\Jp,\Qp)$. This is now a pair of  objectives that depend {\it linearly} on the state frequencies associated with the final augmented state $X_T$.
Accordingly, we define 
$$P_{MQ}=\{(\Jp,\Qp)\mid \pi\in\Pi_{h,u}\}.$$ Note that $P_{MQ}$
is a polyhedron, because it is the image of the polyhedron $Z(\Pi_{h,u})$ under the linear mapping specified by the left-hand sides of Eqs.\ \eqref{eq:a}-\eqref{eq:b}. In contrast,
$P_{MV}$ is the image of $P_{MQ}$ under a nonlinear mapping: 
$$P_{MV}=\{(\lambda,q-\lambda^2)\mid (\lambda,q)\in P_{MQ}\},$$
and is not, in general, a polyhedron.

As a corollary of the above discussion, and for the case of integer rewards, we can exploit convexity to devise pseudopolynomial algorithms for problems that can be formulated in terms of the convex set $P_{MQ}$.
On the other hand, because of the non-convexity of $P_{MV}$, we have not been able to devise pseudopolynomial time algorithms for the problem \mv($\Pi_{h,u}$), or even the simpler problem of deciding whether there exists a policy $\pi\in\Pi_{h,u}$ that satisfies $\Vp\leq v$, for some given number $v$, except for the very special case where $v=0$, which is the subject of our next result. For a general $v$, an approximation algorithm will be presented in the next section.

\begin{theorem}
\begin{itemize}
\item[(a)] If there exists some $\pi\in\Pi_{h,u}$ for which $\Vp=0$, then there exists some $\pi'\in\Pi_{t,s,w}$ for which $V_{\pi'}=0$.
\item[(b)]
Suppose that the immediate rewards are integers, with absolute value bounded by $K$. Then the problem of determining whether there exists a policy $\pi\in\Pi_{h,u}$ for which $\Vp=0$ admits a pseudopolynomial time algorithm.
\end{itemize}
\end{theorem}
\noindent
{\bf Proof.}
\begin{itemize}
\item[(a)]
Suppose that there exists some $\pi\in\Pi_{h,u}$ for which $\Vp=0$. By Theorem \ref{th:equiv}, $\pi$ can be assumed, without loss of generality, to lie in $\Pi_{t,s,w,u}$. Let $\var_{\pi}(W_T\,|\,U_{0:T})$, be the conditional variance of $W_T$, conditioned on the realization of the randomization variables $U_{0:T}$. We have $\var_{\pi}(W_T) \geq \E_{\pi}[\var_{\pi}(W_T\,|\,U_{0:T})]$, which implies that there exists some $u_{0:T}$ such that
$\var_{\pi}(W_T\,|\,U_{0:T}=u_{0:T})=0$. By fixing the randomization variables to this particular $u_{0:T}$, we obtain a deterministic policy, in $\Pi_{t,s,w}$ under which the reward variance is zero.
\item[(b)]
If there exists a policy under which $\Vp=0$, then there exists an integer $k$, with $|k|\leq KT$ such that, under this policy, $W_T$ is guaranteed to be equal to $k$. Thus, we only need to check, for each $k$ in the relevant range, whether there exists a policy such that $(\Jp,\Vp)=(k,0)$. By Theorem \ref{th:pss}, this can be done in pseudopolynomial time. \qed
\end{itemize}

The approach in the proof of part (b) above leads to a short argument, but yields a rather inefficient (albeit pseudopolynomial) algorithm. A much more efficient and simple algorithm is obtained by realizing that the question of whether $W_T$ can be forced to be $k$, with probability 1, is just a reachability game: the decision maker picks the actions and an adversary picks the ensuing transitions and rewards (among those that have positive probability of occurring). The decision maker wins the game if it can guarantee that $W_T=k$. 
Such sequential games
are easy to solve in time polynomial in the number of (augmented) states, decisions, and the time horizon,
by a straightforward backward recursion. 
On the other hand a genuinely polynomial time algorithm does not appear to be possible; indeed, the  proof of Theorem \ref{th:wnpc} shows that the problem is NP-complete.

\section{Approximation Algorithms}
\label{se:approx}
In this section, we deal with the optimization counterparts of the problem \mv($\Pi_{h,u}$).
We are interested in computing approximately the following two functions:
\begin{equation}
v^*(\lambda) = \inf_{\{\pi\in\Pi_{{h,u}}: \Jp\geq \lambda\}} \Vp,
\label{eq:vlambda}
\end{equation}
and
\begin{equation}
\lambda^*(v)  = \sup_{\{\pi\in\Pi_{{h,u}}: \Vp\leq v\}} \Jp.
\label{eq:lambdav}
\end{equation}
If the constraint $\Jp\geq \lambda$ (respectively, $\Vp\leq v$) is infeasible, we use the standard convention $v^*(\lambda)=\infty$
(respectively, $\lambda^*(v)=-\infty$).
Note that the infimum and supremum in the above definitions are both attained, because the set $P_{MV}$ of achievable mean-variance pairs is the image of the polyhedron $P_{MQ}$ under a continuous map, and is therefore compact.

We do not know how to efficiently compute or even generate a uniform approximation of either $v^*(\lambda)$ or $\lambda^*(v)$ (i.e., find a value $v'$ between $v^*(\lambda)-\epsilon$ and $v^*(\lambda)+\epsilon$, and similarly
for $\lambda^*(v)$). In the following two results we consider a weaker notion of approximation that is computable in pseudopolynomial time. We discuss $v^*(\lambda)$ as the issues for $\lambda^*(v)$ are similar.

For any positive $\epsilon$ and $\nu$, we will say that $\hat v (\cdot)$ is an $(\epsilon,\nu)$-aproximation of $v^*(\cdot)$ if, for every $\lambda$,
\begin{equation}\label{eq:app}
v^*(\lambda-\nu)-\epsilon \leq \hat v(\lambda) \leq v^*(\lambda+\nu)+\epsilon.\end{equation}
This is an approximation of the same kind as those considered in \citeA{PY00}: it returns a value $\hat v$ such that $(\lambda, \hat v)$ is an element of the ``$(\epsilon+\nu)$-approximate Pareto boundary'' of the set $P_{MV}$.
For a different view, the graph of the function $\hat v (\cdot)$ is within Hausdorf distance $\epsilon+\nu$ from the graph of the function $v^*(\cdot)$.

We will show how to compute an
$(\epsilon,\nu)$-aproximation in time which is pseudopolynomial, and polynomial in the parameters $1/\epsilon$, and $1/\nu$. 

We start in Section \ref{s:approx1} with the case of integer rewards, and build on the pseudopolynomial time algorithms of the preceding section. We then
consider the case of general rewards in Section \ref{s:approx2}. We finally sketch an alternative algorithm in Section 
\ref{se:approx3} based on set-valued dynamic programming.

\subsection{Integer Rewards}\label{s:approx1}

In this section, we prove the following result.

\begin{theorem} \label{th:approx}
Suppose that the immediate rewards are integers. There exists an algorithm that, given $\epsilon$, $\nu$, and $\lambda$, outputs a value $\hat v(\lambda)$ that satisfies \eqref{eq:app}, and which
runs  in time  polynomial in $|\mS|$, $|\mA|$, $T$, $K$, $1/\epsilon$, and $1/\nu$.
\end{theorem}
\noindent
{\bf Proof.}
Since the rewards are bounded in absolute value by $K$, we have
$v^*(\lambda)=\infty$ for $\lambda>KT$ and
$v^*(\lambda) = v^*(-KT)$ for $\lambda <-KT$. For this reason, we only need to consider $\lambda\in [-KT,KT]$. To simplify the presentation, we assume that $\epsilon=\nu$. We let $\delta$ be such that $\epsilon=3\delta KT$.

The algorithm is as follows. We consider grid points $\lambda_i$ defined by
$\lambda_i=-KT + (i-1)\delta$, $i=1,\ldots,n$, where $n$ is chosen so that
$\lambda_{n-1}\leq KT$, $\lambda_n> KT$. Note that $n=O(KT/\delta)$.
For $i=1,\ldots,n-1$, we calculate $\hat q(\lambda_i)$, the smallest possible value of $\E[W_T^2]$, when $\E[W_T]$ is restricted to lie in $[\lambda_i,\lambda_{i+1}]$. Formally,
$$\hat q(\lambda_i)= \min\Big\{ q \ \big| \  \exists\ \lambda'\in [\lambda_i,\lambda_{i+1}]\ {\rm s.t.}\ 
(\lambda',q)\in P_{MQ}\Big\}.$$
We let $\hat u(\lambda_i)=\hat q(\lambda_i) -\lambda_{i+1}^2$, which can be interpreted as an estimate of the least possible variance when $\E[W_T]$ is restricted to the interval $[\lambda_i,\lambda_{i+1}]$.
Finally, we set
$$\hat v(\lambda)= \min_{i\geq k} \hat u(\lambda_i),\qquad {\rm if}\ \lambda \in [\lambda_k,\lambda_{k+1}].$$ 

The main computational effort is in computing $\hat q(\lambda_i)$ for every $i$. Since $P_{MQ}$ is a polyhedron, this amounts to solving $O(KT/\delta)$ linear programming problems. Thus, the running time of the algorithm has the claimed properties.

We now prove correctness. Let $q^*(\lambda)=\min\{q\mid (\lambda,q)\in P_{MQ}\}$,
and $u^*(\lambda)=q^*(\lambda)-\lambda^2$, which is the least possible variance for a given value of $\lambda$. Note that $v^*(\lambda)=\min\{u^*(\lambda') \mid \lambda'\geq \lambda\}$.

We have $\hat q(\lambda_i) \leq q^*(\lambda')$, for all $\lambda'\in [\lambda_i,\lambda_{i+1}]$. Also, $-\lambda_{i+1}^2 \leq -(\lambda')^2$, for all $\lambda'\in [\lambda_i,\lambda_{i+1}]$. By adding these two inequalities, we obtain $\hat u(\lambda_i) \leq u^*(\lambda')$, for 
all $\lambda'\in [\lambda_i,\lambda_{i+1}]$.
Given some $\lambda$, let $k$ be such that $\lambda\in [\lambda_k,\lambda_{k+1}]$. Then,
$$\hat v(\lambda) =
\min_{i\geq k} \hat u(\lambda_i) \leq \min_{\lambda'\geq \lambda_k}
u^*(\lambda') \leq \min_{\lambda'\geq \lambda} u^*(\lambda')=
v^*(\lambda'),$$
so that $\hat v(\lambda)$ is always an underestimate of $v^*(\lambda)$.

We now prove a reverse inequality.
Fix some $\lambda$ and let $k$ be such that $\lambda\in [\lambda_k,\lambda_{k+1}]$. Let $i\geq k$ be such that $\hat v(\lambda)=\hat u(\lambda_i)$. 
Let also $\bar \lambda\in [\lambda_i,\lambda_{i+1}]$ be such that
$q^*(\bar\lambda) =\hat q(\lambda_i)$.
Note that 
\begin{equation}\label{eq:dl}
\lambda_{i+1}^2- \bar\lambda^2
\leq \lambda_{i+1}^2-\lambda_i^2 
=\delta(\lambda_i+\lambda_{i+1})
\leq 2 \delta( KT+\delta) \leq 3\delta KT.
\end{equation}
Then,
\begin{eqnarray*}
\hat v(\lambda) \stackrel{(a)}{=}\hat u(\lambda_i) &\stackrel{(b)}{=}& \hat q(\lambda_i)-\lambda_{i+1}^2
\stackrel{(c)}{=}q^*(\bar\lambda)-\lambda_{i+1}^2 
\stackrel{(d)}{\geq} q^*(\bar\lambda)- \bar\lambda^2 -3 \delta KT\\
&\stackrel{(e)}{=}&u^*(\bar\lambda) -3 \delta KT
\stackrel{(f)}{\geq} v^*(\bar\lambda)-3 \delta KT
\stackrel{(g)}{\geq} v^*(\lambda-\delta)-3 \delta KT\\
&\stackrel{(h)}{\geq}& v^*(\lambda-\epsilon)-\epsilon.
\end{eqnarray*}
In the above, (a) holds by the definition of $i$; (b) by the definition of $\hat u(\lambda_i)$; (c) by the definition of $\bar\lambda$; and (d) follows from Eq.\ \eqref{eq:dl}. Equality (e) follows from the definition of $u^*(\cdot)$.
Inequality (f) follows from the definition of $v^*(\cdot)$; and (g) is obtained because $v^*(\cdot)$ is nondecreasing and because $\bar\lambda \geq \lambda-\delta$. (The latter fact is seen as follows:
(i) if $i>k$, then $\lambda\leq \lambda_{k+1} \leq \lambda_i \leq \bar\lambda$; (ii) if $i=k$, then both $\lambda$ and $\bar\lambda$ belong to $[\lambda_k,\lambda_{k+1}]$, and their difference is at most $\delta$.) Inequality (h) is obtained because of the definition $\epsilon=3\delta KT$, the observation $\delta<\epsilon$, and the monotonicity of $v^*(\cdot)$.
\qed

\subsection{General Rewards}\label{s:approx2}
When rewards are arbitrary, we can discretize the rewards and obtain a new MDP. The new MDP is equivalent to one with integer rewards to which the algorithm of the preceding subsection can be applied. 
This is a legitimate approximation algorithm for the original problem because, as we will show shortly, the function $v^*(\cdot)$ changes very little when we discretize using a fine enough discretization.

We are given an original MDP $\mM=(T,\mS,\mA,\mR,p,g)$ in which the rewards
are rational numbers in the interval $[-K,K]$, and an approximation parameter
$\epsilon$. We fix
a positive number $\delta$, a discretization parameter whose value will be specified later. We then construct a new MDP $\mM' = ( T,\mS,\mA,\mR',p,g')$, in which the rewards are rounded down to an integer multiple of $\delta$.
More precisely, all 
elements of the reward range $\mR'$ are integer multiples of $\delta$, and for every $t,s,a\in \{0,1,\ldots, T-1\}\times \mS\times \mA$, and any integer $n$, we have
$$
 g_t(\delta n \mid s,a) = \sum_{r:\ \delta n \le r < \delta (n+1)} g_t(r \mid s,a).
 $$
We denote  
 by $J$, $Q$ and by $J'$, $Q'$ the first and second moments of the total reward in the original and new  MDPs, respectively. 
Let $\Pi_{h,u}$ and $\Pi'_{h,u}$ be the sets of (randomized, history-based) policies in $\mM$ and $\mM'$, respectively.   Let $P_{MQ}$ and $P'_{MQ}$ be the associated polyhedra.

We want to to argue that the mean-variance tradeoff curves for the two MDPs are close to each other. This is not entirely straightforward because the augmented state spaces (which include the possible values of the cumulative rewards $W_t$) are different for the two problems and, therefore, the sets of policies are also different. A conceptually simple but somewhat tedious approach involves 
an argument along the lines of \citeA{Whitt78,Whitt79}, generalized to the case of constrained MDPs; we outline such an argument in Section \ref{se:approx3}. Here, we follow an alternative approach, based on a coupling argument.

\begin{proposition}\label{pr:disc}
There exists a polynomial function $c(K,T)$ such that the Hausdorf distance between $P_{MQ}$ and $P'_{MQ}$ is bounded above by $2KT^2\delta$. More precisely,
\begin{itemize}
\item[(a)] 
For every policy $\pi\in \Pi_{h,u}$, there exists a policy $\pi'\in \Pi'_{h,u}$ such that 
$$\max\Big\{ |J'_{\pi'} - J_{\pi}|,\ |Q'_{\pi'} - Q_{\pi}|  \Big\} \leq 2KT^2\delta.$$ 
\item[(b)]
Conversely, for every policy $\Pi'_{h,u}$, there exists a policy $\Pi_{h,u}$ such that
the above inequality again holds.
\end{itemize}
\end{proposition}
\noindent
{\bf Proof.} 
We denote by $d(r)$ the discretized value of a reward $r$, that is,
$d(r)=\max\{ n\delta: n\delta \leq r, \ n\in \Z\}$.
Let us consider a third MDP $\mM''$ which is identical to $\mM'$, except that
its rewards $R''_t$ are generated as follows. (We follow the convention of using a single or double prime to indicate variables associated with $\mM'$ or $\mM''$, respectively.) A random variable $R_t$ is generated according to the distribution prescribed by $g_t(r\,|\, s_t,a_t)$, and its value is observed by the decision maker, who then
incurs the reward $R''_t=d(R_t)$. 
Let $P''_{MQ}$ be the polyhedron associated with $\mM''$.
We claim that $P''_{MQ}=P'_{MQ}$. The only difference between $\mM'$ and $\mM''$ is  that the decision maker in $\mM''$ has access to the additional information $R_t-d(R_t)$. However, this information is incosequential: it does not affect the future transition probabilities or reward distributions.
Thus, $R_t-d(R_t)$ can only be useful as an additional randomization variable. Since $P'_{MQ}$ is the set of achievable pairs using general (history-based randomized) policies, having available an additional randomization variable does not change the polyhedron, and $P''_{MQ}=P'_{MQ}$. Thus, to complete the proof it suffices to show that the polyhedra $P_{MQ}$ and $P''_{MQ}$ are close.

Let us compare the MDPs $\mM$ and $\mM''$. The information available to the decision maker is the same for these two MDPs (since all the history of reward truncations $\{R_\tau - d(R_\tau)\}_{\tau=1}^{t-1}$ is available in $\mM''$ for the decision at time $t$). Therefore, for every policy in one MDP, there exists a policy for the other under which (if we define the two MDPs on a common probability space, involving common random generators) the exact same sequence of states $(S_t=S'_t$), actions $(A_t=A'_t$), and random variables $R_t$ is realized. The only difference is that the rewards are $R_t$ and $d(R_t)$, in $\mM$ and $\mM''$, respectively. 
Recall that $0\leq R_t-d(R_t)\leq \delta$.
We obtain that for every policy $\pi\in\Pi$, there exists a policy $\pi''\in\Pi''$ for which
$0\leq W_T-W''_T = \sum_{\tau=0}^{T-1}
\big(R_t-d(R_t)) \leq \delta T$, and therefore,
$|W_T^2-(W''_T)^2|\leq 2KT^2\delta$.  
Taking expectations, we obtain $|J_{\pi}-J''_{\pi}|\leq T\delta$,
$|Q_{\pi}-Q''_{\pi}|\leq 2KT^2\delta$.
This completes the proof of part (a). The proof of part (b) is identical.
\qed

\begin{theorem} \label{th:approx-gen}
There exists an algorithm that, given $\epsilon$, $\nu$, and $\lambda$, outputs a value $\hat v(\lambda)$ that satisfies \eqref{eq:app}, and which
runs  in time  polynomial in $|\mS|$, $|\mA|$, $T$, $K$, $1/\epsilon$, and $1/\nu$.
\end{theorem}
\noindent
{\bf Proof.}
Assume for simplicity that $\nu=\epsilon$. Given the value of $\epsilon$, let $\delta$ be such that $\epsilon/2= 2KT^2\delta$, and construct the discretized MDP $\mM'$. Run the algorithm from Theorem \ref{th:approx} to find an 
$(\epsilon/2,\epsilon/2)$-approximation $\hat v$ for $\mM'$. Using Proposition~\ref{pr:disc}, it is not hard to verify that this yields an $(\epsilon,\epsilon)$-approximation of $v^*(\lambda)$.
\qed

\subsection{An Exact Algorithm and its Approximation}
\label{se:approx3}

There are two general approaches for constructing approximation algorithms. (i) One can discretize the problem, to obtain an easier one, and then apply an algorithm specially tailored to the discretized problem; this was the approach in the preceding subsection. (ii) One can design an exact (but inefficient) algorithm for the original problem and then implement the algorithm approximately. This approach will work provided the approximations do not build up excessively in the course of the algorithm. In this subsection, we elaborate on the latter approach.

We defined earlier the polyhedron $P_{MQ}$ as the set of achievable first and second moments of the cumulative reward starting at time zero at the initial state. We extend this definition by considering intermediate times and arbitrary (intermediate) augmented states. 
We let
\begin{eqnarray}
C_t(s,w) = \big\{ (\lambda,q): \exists \pi\in \Pi_{h,u} \ {\rm s.t.}&  \E_\pi [W_T \mid  S_t = s,\, W_t=w] =\lambda\mbox{ and } \label{eq:Cdef}\\
&  \E_\pi [W_T^2\mid S_t = s,\, W_t=w)=q\big\}.\nonumber
\end{eqnarray}
Clearly,  $C_0(s,0)=P_{MQ}$. 
Using a straightforward backwards induction, it can be shown
that $C_t(\cdot,\cdot)$ satisfies the set-valued dynamic programming recursion
\footnote{If $X$ and $Y$ are subsets of a vector space and $\alpha$ a scalar, we let $\alpha X =\{\alpha x \mid x\in X\}$
and $X+Y=\{x+y\mid x\in X,\ y\in Y\}$. Furthermore, if for every $a\in \mA$, we have a set $X_{\alpha}$, then ${\rm conv}_{a\in\mA}\{X_{\alpha}\}$ is the convex hull of the union of these sets.}
\begin{equation}
C_t(s,w)  =  \conv_{a\in \mathcal{A}}\left\{ \sum_{s'\in \mathcal{S}} p_t (s'\mid s,a) \sum_{r\in \mathcal{R}} g_t(r \mid s,a) C_{t+1}(s',w+r)\right\},\label{eq:recursion}
\end{equation}
for every $s\in \mathcal{S}$, $w\in\R$, and for $t=0,1,2,\ldots, T-1$, 
initialized with the boundary conditions
\begin{equation}\label{eq:boundary}
C_T(s,w) =  \{(w,w^2)\}.
\end{equation}

A simple inductive proof shows that the sets $C_t(s,w)$ are polyhedra; this is because $C_T(s,w)$ is either empty or a singleton and because the sum or convex hull of finitely many polyhedra is a polyhedron. Thus, the recursion involves a finite amount of computation, e.g., by representing each polyhedron in terms of its finitely many extreme points. In the worst case, this translates into an exponential time algorithm, because of the possibly large number of extreme points. However, such an algorithm can also be implemented approximately. If we allow for the introduction of an $O(\epsilon/T)$ error at each stage (where error is measured in terms of the Hausdorf distance), we can work with approximating polyhedra that involve only $O(1/\epsilon)$ extreme points, while ending up
with a $O(\epsilon)$ total error; this is because we are approximating polyhedra in the plane, as opposed to higher dimensions where the dependence on $\epsilon$ would have been worse dependence. The details are straightforward but somewhat tedious and are omitted.
On the other hand, in practice, this approach is likely to be faster than the algorithm of the preceding subsection.

\section{Conclusions}
\label{s:concl}

We have shown that mean-variance optimization problems for MDPs are typically NP-hard, but sometimes admit  pseudopolynomial approximation algorithms.
We only considered finite horizon problems, but it is clear that the negative results carry over to their infinite horizon counterparts. Furthermore,
given that the contribution of the tail of the time horizon in infinite horizon discounted problems (or in ``proper'' stochastic shortest path problems as in \citeA{Bertsekas95}) can be made arbitrarily small, our approximation algorithms can also yield approximation algorithms for infinite horizon problems.

Two more problems of some interest deal with finding a policy that has the smallest possible, or the largest possible variance. There is not much we can say here, except for the following:
\begin{itemize}
\item[(a)] The smallest possible variance is attained by a deterministic policy, that is, 
$$\min_{\pi\in \Pi_{{h,u}}} V_\pi = \min_{\pi\in \Pi_{{h}}} V_\pi.$$ This is proved using the inequality $\var_{\pi}(W_T) \geq \E_{\pi}[\var_{\pi}(W_T\,|\,U_{0:T})]$.
\item[(b)] Variance will be maximized, in general, by a randomized policy. To see this, consider a single stage problem and two actions with deterministic rewards, equal to 0 and 1, respectively. Variance is maximized by assigning probability 1/2 to each of the actions.
The variance maximization problem is equivalent to maximizing the concave function $q-\lambda^2$ subject to $(\lambda,q)\in P_{MQ}$. This is a quadratic programming problem over the  polyhedron $P_{MQ}$ and therefore admits a pseudopolynomial time algorithm, when the rewards are integer.
\end{itemize}

Our results suggest several interesting directions for future research, which we briefly outline below.

First, our negative results apply to general MDPs. 
It would be interesting to determine whether the hardness results remain valid for specially structured MDPs. One possibly interesting special case involves multi-armed bandit problems:
there are $n$ separate MDPs (``arms"); at each time step, the decision maker has to decide which MDP to activate,
while the other MDPs remain inactive. Of particular interest here are index policies that compute a value (``index") for each MDP and select an MDP with maximal index; such 
policies are often optimal for the classical formulations (see \citeA{Gittines79} and \citeA{Whittle88}). 
Obtaining a policy that uses some sort of an index for the mean-variance problem 
or alternatively proving that such a policy cannot exist would be  interesting.

Second, a number of complexity questions have been left open.
We list a few of them:
\begin{itemize}
\item[(a)] Is there a pseudopolynomial time algorithm for computing $v^*(\lambda)$ or $\lambda^*(v)$ exactly?
\item[(b)] Is there a polynomial or pseudopolynomial time algorithm that computes $v^*(\lambda)$ or $\lambda^*(v)$ within a uniform error bound $\epsilon$?
\item[(c)] Is the problem of computing $\hat v(\lambda)$ with the properties in Eq.\ \eqref{eq:app} NP-hard?
\item[(d)] Is there a pseudopolynomial time algorithm the smallest possible variance in the absence of any constraints on the mean cumulative reward?
\end{itemize}


Third, bias-variance tradeoffs may pay an important role in speeding up certain control and learning heuristics, such as those involving control variates \cite{Meyn08book}. 
Perhaps mean-variance optimization can be used to address the exploration/exploitation tradeoff in model-based reinforcement learning, with variance reduction serving as a means
 to reduce the exploration time (see \citeA{SutBar98} for a general discussion of exploration-exploitation in reinforcement learning). 
Of course, in light of the computational complexity of bias-variance tradeoffs, incorporating bias-variance tradeoffs in 
learning makes sense only if experimentation is nearly prohibitive and computation time is cheap.
Such an approach
could be particularly useful if a coarse, low-complexity,  approximate solution of a bias-variance tradeoff problem can result in significant exploration speedup.

Fourth, we only considered mean-variance tradeoffs in this paper. However, there are other interesting and potentially useful criteria that can be used to incorporate risk into multi-stage decision making. For example, 
\citeA{LiuKoenig05:risksensitiveoneswitch} consider a utility function with a single switch. Many other risk aware criteria have been considered in the single stage case. It would be interesting to develop a comprehensive theory for the complexity of solving multi-stage decision problems under general (monotone convex or concave) utility function and under risk constraints. This is especially interesting for the approximation algorithms presented in Section \ref{se:approx}.

\subsubsection*{Acknowledgments}{This research was partially supported by the Israel Science Foundation (contract 890015), a Horev Fellowship, and the National Science Foundation under grant CMMI-0856063.} 

\bibliographystyle{apacite}
\bibliography{qmdp}

\end{document}